\documentclass[journal]{IEEEtran}
\usepackage{amsmath,amsfonts}
\usepackage{algorithmic}
\usepackage{algorithm}
\usepackage{array}
\usepackage[caption=false,font=normalsize,labelfont=sf,textfont=sf]{subfig}
\usepackage{textcomp}
\usepackage{stfloats}
\usepackage{url}
\usepackage{verbatim}
\usepackage{graphicx}
\usepackage{cite}
\usepackage{xcolor}
\usepackage{multirow}
\usepackage{makecell}
\usepackage{amsmath}
\usepackage{amssymb}

\usepackage{xcolor}

\usepackage[switch]{lineno}

\begin{document}

\title{Graph Neural Networks for Cross-Camera Data Association}
 \author{Elena Luna, Juan C. SanMiguel, Jos\'{e} M. Mart\'{i}nez, and Pablo Carballeira
 \thanks{Elena Luna, Juan C. SanMiguel, Jos\'{e} M. Mart\'{i}nez, and Pablo Carballeira are with the Video Processing and Understanding Lab, Universidad Aut\'{o}noma de Madrid, Spain, e-mail: \{elena.luna;~juancarlos.sanmiguel;~josem.martinez;~pablo.carballeira\}@uam.es.}}
 

\maketitle

\begin{abstract}
Cross-camera image data association is essential for many multi-camera computer vision tasks, such as multi-camera pedestrian detection, multi-camera multi-target tracking, 3D pose estimation, etc. 
This association task is typically stated as a bipartite graph matching problem and often solved by applying minimum-cost flow techniques, which may be computationally inefficient with large data. Furthermore, cameras are usually treated by pairs, obtaining local solutions, rather than finding a global solution at once. Other key issue is that of the affinity measurement: the widespread usage of non-learnable pre-defined distances, such as the Euclidean and Cosine ones. This paper proposes an efficient approach for cross-cameras data-association focused on a global solution, instead of processing cameras by pairs. To avoid the usage of fixed distances, we leverage the connectivity of Graph Neural Networks, previously unused in this scope, using a Message Passing Network to jointly learn features and similarity. We validate the proposal for pedestrian multi-view association, showing results over the EPFL multi-camera pedestrian dataset. Our approach considerably outperforms the literature data association techniques, without requiring to be trained in the same scenario in which it is tested. Our code is available at \url{http://www-vpu.eps.uam.es/publications/gnn_cca}.
\end{abstract}


\section{Introduction}\label{sec:intro}
\IEEEPARstart{D}{ata} association across camera views is an intermediate key step in many complex computer vision tasks, such as multi-camera 3D pedestrian detection \cite{lima2021generalizable}, 3D pose estimation for multiple views \cite{dong2019fast,chen2020cross,kadkhodamohammadi2021generalizable}, multi-view multi-target tracking \cite{leal2012branch,zhang2008global,he2020multi}, and even robotic perception \cite{aragues2011consistent}, among others. As the end result of all these tasks depends, to a large extent, on how good the association is, it is worth investing in finding an optimal solution.

 This association is typically stated as a bipartite graph matching problem and solved applying minimum-cost flow techniques, e.g., resolving an association matrix using the Hungarian algorithm \cite{chu2019online, tan2019multi,xu2019spatial}. Re-Identification appearance cues \cite{he2020city} that may be combined with 3D spatial location \cite{chen2019multi,qian2020electricity} are widely used to build the association matrix. However, due to their computational cost, the use of minimum-cost flow-based solvers in practical implementations is limited to datasets not containing large number of data \cite{wang2019efficient}.
 
A fundamental issue in data association is the affinity measurement, i.e., the likelihood that two objects from different views belong to the same instance. In order to compare location coordinates (on image or gound-plane) and appearance features of the objects, the Euclidean and Cosine distances are widely used \cite{Narayan2017,ristani2018features,Maksai_2019_CVPR,he2020multi,zhou2021learning}. 
However, it is an unsupervised and sub-optimal approach \cite{gou2018systematic} since these fixed metrics are not adapted to the target domain. Moreover, a manual threshold is often required to take an association decision. 
Metric learning \cite{kulis2012metric} incorporates training data in a supervised manner to achieve a better performance. It aims to learn a task-specific distance function, thus, a new feature space, such that features of the same instance are close.

Other challenge is to handle the process of $n$-view matching (i.e., $n>2$). Previous works have been limited to perform cross-camera data association by pair of views \cite{dong2019fast, leal2012branch,kadkhodamohammadi2021generalizable}. However, matching each pair of views independently may produce undesired cycle-inconsistent correspondences \cite{phillips2019all}, i.e., a pair of corresponding data in two different views not corresponding to the same data in another view.

Graphs are a type of structure that model data as a set of objects or concepts (represented by nodes) and the relationships between them (edges). Recently, traditional graph analysis using machine learning techniques has been applied to many fields such as social sciences \cite{hamilton2017inductive,kipf2016semi}, natural sciences \cite{sanchez2018graph,NIPS2017_f5077839}, knowledge graphs \cite{ijcai2017-250}, and computer vision \cite{liang2016semantic,hu2018relation,sheng2018heterogeneous}, among other research fields. Graph Neural Networks (GNNs) were initially introduced by \cite{scarselli2008graph} as an adaptation of Neural Networks (NNs) to work on graph structures, including cyclic, directed, and undirected graphs. GNNs are based on an information diffusion mechanism. The main idea is that a graph is built as a set of nodes linked according to the graph connectivity. The nodes and edges may update their hidden states by exchanging (propagating), information along their neighbourhood (adjacent nodes). Finally, the output is computed based on the states, locally at node or edge-level, or globally at graph-level, depending on the target task. The information (message) passing mechanisms along the graph structure were reformulated as a common framework called Message Passing Networks (MPNs) by \cite{gilmer2017neural} and later extended in \cite{battaglia2018relational}.

In this paper, we propose to treat the cross-camera data association task as a learning problem. We learn, simultaneously, the feature representations and the similarity function to merge bounding box detections of the same object from different camera views, aiming to deal with the view-point variation challenge.  The idea of learning both the feature representation as well as the similarity at the same time has been already proposed and its effectiveness proven in other scopes such as single-view multi object tracking \cite{braso2020learning}, human pose recovery \cite{kanazawa2018end}, and vehicle re-identification \cite{zhu2018joint}, among others; however, to the best of our knowledge, it has not yet been considered for the cross-camera data association task. To this aim we propose to learn a solution by training a GNN, since it has been proven effective in other similar tasks,  being, as far as we know,  the first paper tackling the cross-camera data association task using GNNs. We leverage the appearance and spatial information without making any prior assumption about the true number of objects in the scene. We perform learning directly in the graph domain with a Message Passing Network, thus providing a global association solution for all the cameras at once, instead of performing association by pairs. 

In summary, the main contributions of this paper include:
\begin{itemize}
    \item A new approach fully devoted to associate data from different views frame by frame, giving a global solution at once, instead of finding local sub-optimal solutions.
    \item The use of GNNs to solve the cross-camera data association task, previously unused in this scope.
    \item A novel approximation based on similarity learning, yet unconsidered for this task, to avoid the usage of fixed thresholded distances.
    \item An extensive ablation study and comparison with state of the art techniques showing the favorable performance of the proposed approach (we reach an improvement of 18.55\% over the better state-of-the-art approach).
    
\end{itemize}

\begin{figure*} \label{fig:block-diagram}
    \centering
    \includegraphics[width=0.99\textwidth,keepaspectratio]{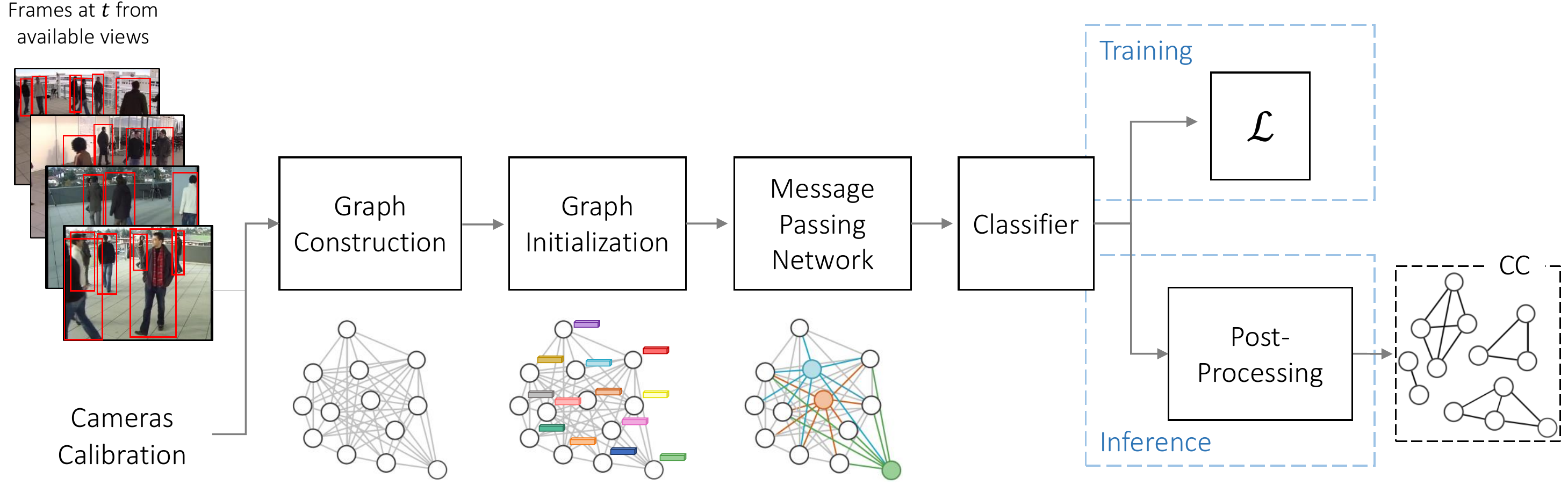}
    \caption{Block diagram of the proposed approach. The Graph Construction module creates the graphs and defines the connections between nodes. In Graph Initialization, an initial feature embedding is associated to each node and edge. The Message Passing Network propagates and combines the embeddings of the nodes and edges along the graph. Finally, the Classifier defines which edges in the graph are active. When training, the loss function is computed and backpropagated to perform edge classification. When inference, the graph is pruned according to each edge's likelihood in the Post-Processing stage. Finally, the Connected Components of the graph are computed to define the association between detections of different views. }
\end{figure*}

\section{Related Work}
Within the computer vision field, cross-camera data association is the process of finding the correspondences between the data, typically points, regions or bounding boxes, of different camera views. This work focuses on associating multi-view detections frame by frame, and processing all the camera views global and simultaneously.

%

 \subsection{Multi-view Association}
The traditional multi-view data association approaches address the issue by finding local optimum solutions, opposed to finding a global solution. Thus, solving a sequence of decoupled pairwise matching sub-tasks, i.e. by pairs of cameras independently. Moreover, these matchings tend to be solved as linear assignment problems. This greedy approximation does not leverage the redundancy in the data and often leads to cycle-inconsistent associations \cite{fathian2020clear}. For instance, a detection in view 1 is matched to another in view 2, and to another in view 3, but the detections in views 2 and 3 are not matched together. A multi-view multi-target tracking approach performing cross-camera data association solving a min-cost problem over conventional graphs exploiting spatial and temporal information was presented in \cite{leal2012branch}. However, they proposed association only between pairs of cameras.
 The task of multi-person 3D pose estimation from multiple views was tackled in \cite{dong2019fast} by carrying out data association across views limited to a pair of cameras at the same time.

\subsection{Affinity Measure}

Since deep learning spread to computer vision (e.g., in re-identification field \cite{yi2014deep, li2014deepreid}), deep features (i.e., embeddings) are chosen to describe bounding boxes' appearance for multi-view matching. In order to measure the affinity between these appearance features, the use of non-learnable pre-defined distances, such as Euclidean and Cosine distances, was widely spread \cite{Narayan2017,ristani2018features,Maksai_2019_CVPR,he2020multi,zhou2021learning}. 
The Euclidean distance between spatial  features is also used by \cite{lopez2018semantic} to associate multi-camera people detection. An extension of this work was proposed by \cite{lima2021generalizable}, in which they also use the Euclidean distance between location and appearance features to associate detections. In \cite{dong2019fast}, they combine re-identification appearance and geometrical information performing a linear assignment based on the Euclidean distance for the similarity computation, for the task of multi-camera 3D pose estimation. Although the usage of this fixed distances is broadly common, it has been proven to be less efficient than similarity learning \cite{koestinger2012large,liao2015person,guillaumin2009you,davis2007information}. Similarity learning involves three main processes: 1) transformation of the data into features vectors using a deep learning architecture (typically an encoder); 2) pairwise comparison of the vectors by using a distance metric (typically Euclidean); and 3) classification of this distance as similar or dissimilar (using any classifier). Since we adopt this methodology, these three concepts are present in our approach.
    
\subsection{Graph Neural Networks}
Graph Neural Networks are used in several scopes, however they have not been exploited until recently in computer vision. GNNs have been applied in point feature matching \cite{phillips2019all, fathian2020clear}, gesture learning \cite{xie2021sequential}, video moment retrieval \cite{gao2021learning}, visual question answering \cite{narasimhan2018out} or single-camera single-object tracking \cite{gao2019graph}. Regarding single-camera multi-object tracking,  \cite{Papakis2020} proposes the use of GNN to extract node and edge embeddings, but computing similarity using the cosine distance and perform data association by using a linear assignment, i.e., Hungarian Algorithm. The first approach of performing feature and similarity learning jointly for associating detections was introduced in \cite{braso2020learning} by proposing a time-aware MPN variation: detecting associations across time to perform batch-based/offline single-camera multi-object tracking. Although they propose a very novel and well-developed idea, the appearance variation of a detected object (bounding box) from frame to frame is not very significant, in contrast to multi-view association, where the view-point variation is the main challenge. 

\section{Method}
Although the proposed approach can be applied to associate any type of cross-camera object detections, we focus our evaluation on pedestrians. We exploit the connectivity of a Graph Neural Network (GNN) using a Message Passing Network (MPN)  \cite{gilmer2017neural, battaglia2018relational} to perform feature learning as well as to provide a solution for cross-camera association by computing edge classification.  It is composed of the following main stages: Graph Construction (Section \ref{sec:graph-const}), Graph Initialization (Section \ref{sec:feat-enc}), Message Passing Network (Section \ref{sec:mpn}) and the Classifier (Section \ref{classifier}). The Post-Processing stage (Section \ref{seq:post}) and the computation of the Connected Components (Section \ref{sec:CC}) are only performed when inference.

\subsection{Problem Formulation \label{sec:prob-form}} 

Let us define a graph $G=(V,E)$, where $V$ is a set of nodes and $E$ is the set of edges between them. On the basis of GNNs, we can give each unit in a graph, i.e. nodes and edges, a state representation, i.e., embedding, to represent its concept. In our case, nodes stand for all pedestrian detections in the available views, and edges for the relationship between them.  The initialization, update and classification of nodes, and edges embeddings are described in the following subsections. 

Let us assume $\mathcal{C} = \left\{c_{i}, \; i \in [1,M] \right\}$ as a set of $M$ simultaneously available cameras with overlapping fields of view (FoVs).

Let $\mathcal{P}_t=\left\{ p_i, \; i\in [1,|\mathcal{P}_t|\;]\right\}$, be the set of detected pedestrians in a scene at frame $t$.  Each person identity $p_i$ is formed by a set of detections coming from a single or multiple camera views. Equivalently, an identity can be defined as a set of nodes in the graph such that $p_i = \left\{v_{j}, \; j \in [1,M_{p_{i}}] \right\}$, being $M_{p_{i}}$ the number of cameras detecting the person $p_i$.

The goal is, at each frame $t$, to group the nodes in the graph $G_{t}=(V,E)$ corresponding to the same people identity, obtaining as many clusters of nodes as pedestrians appear. To this aim, we perform edge classification based on the final edge embedding. Let us consider a binary variable for each edge in the graph in order to represent the graph partitions. The binary variable $y_{(v_i, v_j)}$ for the edge between the nodes $v_i$ and $v_j$ is defined as:

\begin{equation}
y_{(v_i, v_j)}=\left\{\begin{array}{ll}
1 & \exists \; p_{i} \in \mathcal{P}_{t} \; \textrm{ s.t. } \; (v_i, v_j) \in p_{i} \\
0 & \textrm{ otherwise.}
\end{array}\right.
\end{equation}
In other words, it is defined to be 1 (active) between nodes that belong to the same pedestrian, and 0 (non-active) otherwise.  Thus, each person in the scene can be mapped into a group of nodes in the graph, in other words, a Connected Component(CC) in the graph.

\subsection{Graph Construction} \label{sec:graph-const}

As input we consider, at each frame $t$, a set of pedestrian detections $\mathcal{D}_t = \left\{d_{i}, \; i \in [1,N_t]\right\}$, being $N_t$ the number of total detections at $t$. Each detection is defined by  $d_i = (\textbf{b}_i, c_i)$, where $\textbf{b}_i$ and $c_i$ stand for its bounding box image and camera number, respectively. 

We model the problem with an undirected graph $G_{t}=(V,E)$, where all the edges are bidirectional. Thus, at each frame $t$ a graph $G_t$ is constructed so that all detections from different camera views are connected. We apply this constraint since we work under the assumption that two detections coming from the same camera cannot belong to the same identity (person $p_i$). 

Let $V=\left\{v_{i}, \; i \in [1,N_t]\right\}$ be the set of nodes, and each node $v_i \in V$ denotes a single detection $d_i \in \mathcal{D}_t$. Let $E$ be the set of edges connecting pair of nodes as follows:
 \begin{equation} \label{Eq-1}
 E=\left\{\left(v_{i}, v_{j}\right), c_i \neq c_j \right\}.
 \end{equation}
Note that nodes under the same camera are not connected between them.

\subsection{Graph Initialization: Feature Encoding} \label{sec:feat-enc}

\begin{figure}
    \centering
    \includegraphics[width=0.49\textwidth,keepaspectratio]{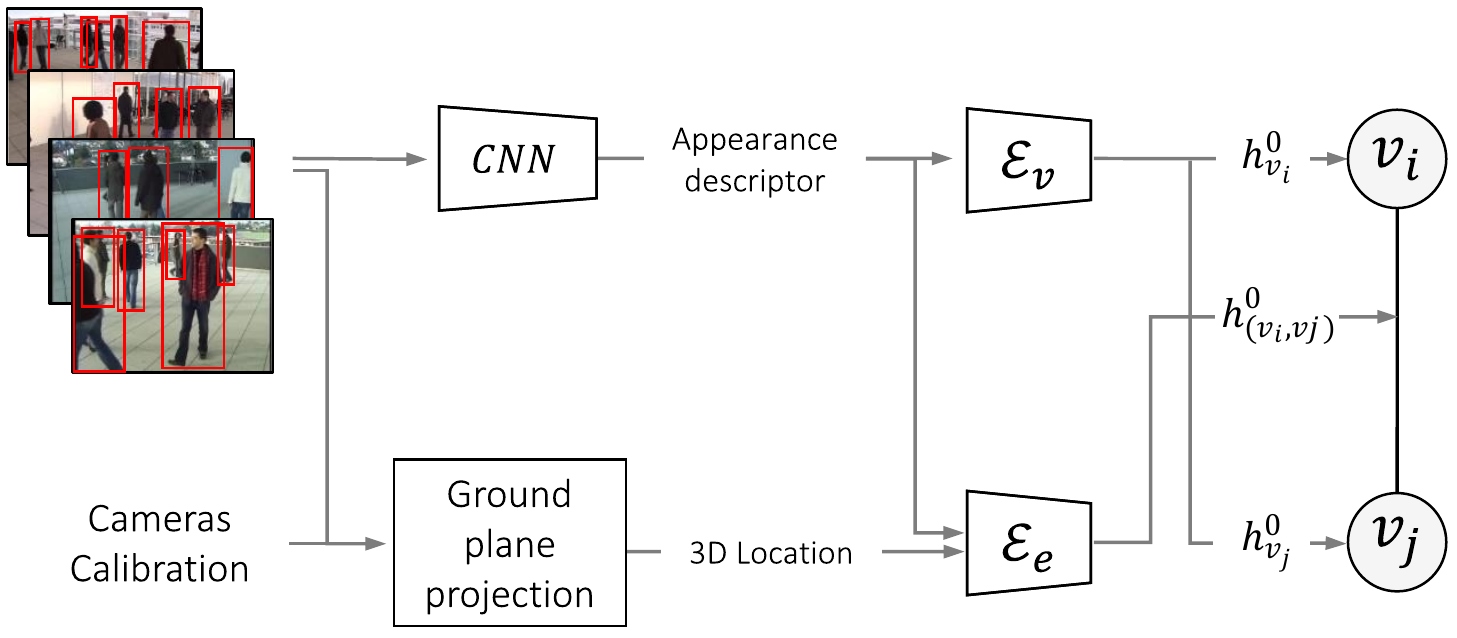}
    \caption{Graph Initialization. \label{fig:graph-ini}}
\end{figure}

As stated in Section \ref{sec:intro}, the target of GNNs is to learn final state embeddings, for each node and edge in the graph, containing information about their neighbourhood, which can be used to generate an output, such as the node or edge prediction label. This section details the states initialization; update and classification are explained in the following subsections. All the nodes and edges in $G_t$ are initialized with their respective initial state embeddings (see Figure \ref{fig:graph-ini}).

The initial nodes embbedings $h_{v_{i}}$ rely only on appearance information. For each detection $d_i \in \mathcal{D}_t$, its bounding box image $\textbf{b}_i$ is fed to a Convolutional Neural Network (CNN) to extract its appearance descriptor.  The latter goes through the learnable node encoder $\mathcal{E}_v$, which outputs the node embedding $h_{v_i}$ used to initialize node $v_i$. It can be defined as follows: 
\begin{equation}
    h_{v_{i}} = \mathcal{E}_{v}( \textrm{\small CNN} (\textbf{b}_i)).
\end{equation}

On the other hand, the initial edges embeddings rely on a concatenation of the appearance and spatial information related to the pair of nodes that the edge is connecting. Let us define the related appearance feature between two nodes as follows:

\begin{equation} 
\begin{split}
\Delta f_{i,j}  = [\; &||\; \textrm{\small CNN}(\textbf{b}_i) ,   \textrm{\small CNN}(\textbf{b}_j)\;||_2, \\
 & \textit{\small cos\_similarity}(\textrm{\small CNN}(\textbf{b}_i),\textrm{\small CNN}(\textbf{b}_j))\;]
\end{split}
\end{equation}

where $[\cdot,\cdot]$ denotes the concatenation of the Euclidean distance and Cosine similarity between the appearance descriptors of a pair of bounding boxes $\textbf{b}_i$ and $\textbf{b}_j$. Both distances are considered, as they are the most commonly used for computing features distances, to obtain a higher dimension and more distinctive feature.

Since we only consider cross-camera associations (Equation \ref{Eq-1}), in order to obtain a relative spatial distance between detections from different cameras, a transformation to a common ground plane is required. More specifically, we need to obtain an approximation, as accurate as possible, of the foot position of each detection in a common framework.
Given a bounding box $\textbf{b}_i$ associated to the node $v_i$, the middle point of its base is defined by $(x_i + \frac{w_i}{2}, y_i)$, being $(x_i, y_i, w_i, h_i)$ the upper-left corner pixel coordinates, the width and height. This point is projected to the common ground plane as
\begin{equation}
     (X_{i}, Y_{i}) = \textbf{H}_{c_i}(x_i + \frac{w_i}{2}, y_i),
\end{equation}
being  $\textbf{H}_{c_i}$ the homography matrix that transforms coordinates from the image plane of the camera $c_i$ to coordinates in the common ground plane. 
Hence, we compute the relative spatial distance information between two nodes as 
\begin{equation} 
\begin{split}
  \Delta s_{i,j}  = [\; &||\; (X_i,Y_i), (Y_j, Y_j)\; ||_1, \\  
   &||\; (X_i,Y_i), (X_j,Y_j)\; ||_2\;],
\end{split}
\end{equation}

the concatenation of the Manhattan and Euclidean distances, as they are the most commonly used for computing spatial distances, in order to obtain a more distinctive feature.

Finally, the concatenation of the distances, $ \Delta f_{i,j}$ and $ \Delta s_{i,j}$,  is fed to a learnable edge encoder $\mathcal{E}_e$ to obtain the initial edge embedding:
\begin{equation}
h_{(v_{i},v_{j})} = \mathcal{E}_{e}( [\Delta f_{i,j},\; \Delta s_{i,j}]).
\end{equation}

Both Multi Layer Perceptron (MLP) encoders (FCs + ReLU), $\mathcal{E}_{v}$ and $\mathcal{E}_{e}$, learn the optimal adjustment of the features to the target association task.


\subsection{Message Passing Network} \label{sec:mpn}

\begin{figure}
    \centering
    \includegraphics[width=0.45\textwidth,keepaspectratio]{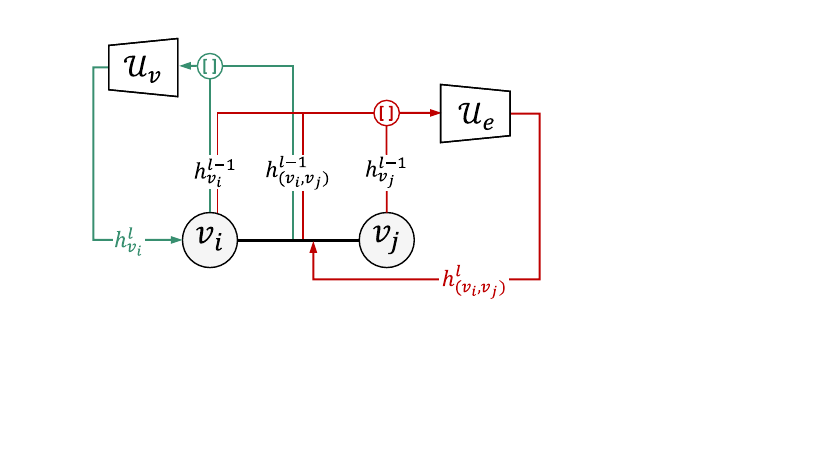}
    \caption{Block diagram describing the edge and node update steps of the MPN at a given step $l$.  \label{fig:mpn}}
\end{figure}


MPNs propagate neural messages between neighbouring nodes and edges.  Formally, each node and edge are initialized with a starting representation, usually called initial state. The last state is commonly used for prediction. Figure 
 \ref{fig:block-diagram} depicts the main blocks diagram of the method.

The goal of MPNs is to propagate neural messages between neighbouring nodes and edges along the graph $G_t$. This diffusion is accomplished by embedding updates, known as message passing steps, introduced by \cite{gilmer2017neural}. At each propagation step, each node's and edge's sent messages are computed; then, every node and edge aggregates their received messages; and, finally, their representation is updated by merging the incoming information with their own previous representation. Formally, each node and edge are initialized with a starting representation, usually called initial state (see Section \ref{sec:feat-enc}). The last state is commonly used for prediction (see Section \ref{classifier}).
Our formulation, is based on \cite{battaglia2018relational, kipf2016semi} and it divides each message passing step in two stages: edge update and node update. Both updates are iteratively performed over $L$ iterations, i.e., $L$ message passing steps. Note that the higher $L$ is, the farther the information is propagated along the graph. It plays a similar role to the receptive field of Convolutional Neural Networks (CNNs). For each iteration $l \in [1,L]$, the embedding of the edge connecting nodes $v_i$ and $v_j$ is updated as follows:
\begin{equation}
 h_{(v_{i},v_{j})} ^{l} = \mathcal{U}_e ([ h_{v_i}^{l-1}, h_{v_j}^{l-1}, h_{(v_i,v_j)}^{l-1}]),
\end{equation}
where $\mathcal{U}_e$ is a learnable MLP encoder (FCs + ReLU). 
On the other hand, the node embedding is updated by aggregating all the messages coming to the node $v_i$ from its adjacent nodes, i.e., its neighbors (see Figure \ref{fig:mpn}):
\begin{equation}
    h_{v_i}^l = \sum_{j\in \mathcal{N}(v_i)} m_{(v_i,v_j)}^l,
\end{equation}
where $\mathcal{N}(\cdot)$ denotes the neighbouring nodes, and 

\begin{equation}
    m_{(v_i,v_j)}^l = \mathcal{U}_v([ h_{v_i}^{l-1}, h_{(v_i,v_j)}^{l}]),
\end{equation}
where $\mathcal{U}_e$ is another learnable MLP encoder (FCs + ReLU). The update functions $\mathcal{U}_e$ and  $\mathcal{U}_v$ are learnt and shared across $G_t$. It is worth to mention that during all the iterations of the message passing procedure, the propagation of information happens simultaneously for all nodes and edges in $G_t$. Also note that during the first iteration $l=1$, $l-1 = 0$ denotes the initial state of the graph, just after graph construction (section \ref{sec:graph-const}) and features initialization (section \ref{sec:feat-enc}).

\subsection{Classifier} \label{classifier}

Our proposal is learning to predict which edges in the graph are active (being the same person identity). We train our model to predict $\hat{y}_{(v_i,v_j)}$, the value of the binary variable $y_{(v_i, v_j)}$ for each edge in $G_t$. It can be seen as an edge classification problem, using $y_{(v_i, v_j)}$ as labels. During inference, the graph is pruned according to each edge's prediction and a post-processing strategy is follow to fulfill some constraints (see Section \ref{seq:post}). 
 
 The classification of a given edge at a given iteration is computed by:
 
 \begin{equation}
      \hat{y}_{(v_i,v_j)}^l = \mathcal{C}(h_{(v_i,v_j)}^l),
 \end{equation}
 being $\mathcal{C}$ the learnable classifier, i.e., another MLP (FC + ReLU) followed by a sigmoid function, that outputs a single prediction value. 
 
 \subsubsection{Training}
 To compute the training loss of the graph $G_t$, we use the classical binary Cross-Entropy loss (CE)  \cite{goodfellow2016deep}, aggregated over all edges in $E$ (equation \ref{Eq-1}) and all iterations, similarly to \cite{braso2020learning}:
 \begin{equation}
     \mathcal{L}_{G_t} = \sum_{l=1}^L \sum_{(v_i,v_j)\in E} \textrm{CE}(\hat{y}^{l}_{(v_i,v_j)},y_{(v_i, v_j)} ).
 \end{equation}
 
 In the end, we learn a method capable to directly predict partitions in the graph by performing edge classification.

\subsubsection{Inference}
As previously mentioned, the goal is, at each frame $t$, to group the nodes in the graph $G_t$ corresponding to the same person identity, obtaining as many CCs as pedestrians appear.
To infer these components in the graph, we consider the output of the trained MPN model at the final iteration $l=L$. Thus, we obtain, for each edge in $G_t$, $\hat{y}_{(v_i,v_j)}^L \in [0,1]$, denoting the probability of the edge of being active. Then, this likelihood prediction is binarized, as usual in the literature, as follows:
\begin{equation}
    \hat{y}_{(v_i,v_j)}^B=
    \begin{cases}
      0, & \text{if}\ \hat{y}_{(v_i,v_j)}^L \; < 0.5 \\
      1, & \text{otherwise.}
    \end{cases}
  \end{equation}
  
$\hat{y}_{(v_i,v_j)}^B$ is used to classify edges as actives or non-actives. Non-active edges are prunned, while the active ones are maintained. The last stage of the inference is the Post-processing strategy, that is described hereunder.

\subsection{Post-processing \label{seq:post}} 
\begin{figure} 
    \centering
    \includegraphics[height=9cm,keepaspectratio]{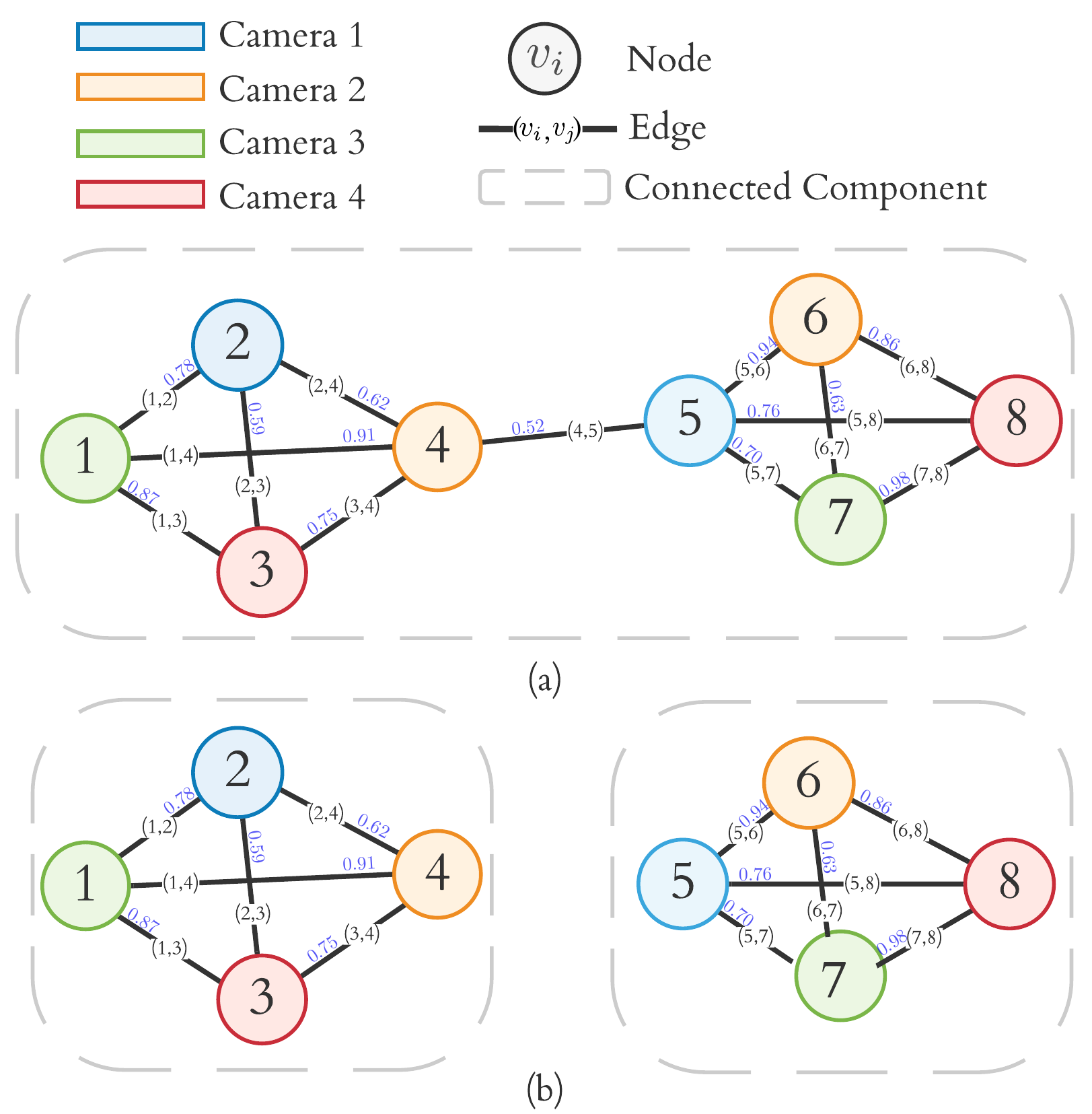}
    \caption{\textit{Pruning} Post-processing example of a graph with $M=4$ cameras. (a) shows the initial state of the graph after inference and before Post-processing stage. Nodes 4 and 5 are not fulfilling the condition in equation \ref{eq:3} since they are connected to more than $M-1=3$ cameras. As both nodes have only one bridge edge between the edges under consideration to remove, the bridge edge (4,5) is removed and the final correct sub-graphs are obtained, see (b). \label{fig:FlowConstr1}}
\end{figure}

Let us define the flow of a node as the number of edges connected to it. As each node can only belong to a single identity, we work under the assumption that each node can be connected to, at most, $M-1$ nodes, thus each pedestrian identity can be composed of, at most, detections of $M$ views.
Therefore, for each node $v_i$ the first following flow constraint must be satisfied:

\begin{equation} \label{eq:3}
    \sum^{V}_{j=1} \hat{y}_{(v_i,v_j)}^B \leq (M-1), \; \forall j \neq i,
\end{equation}
meaning that each node can be connected to, at most, other $M-1$ nodes, being $M$ the number of total cameras (see figure \ref{fig:FlowConstr1}). 
Graph theory \cite{biggs1986graph} defines \textit{cycles} in graphs as non-empty closed paths starting and ending at the same node. On the other hand, \textit{bridges} in a graph are edges not contained in any cycle, i.e., whose suppression increases the number of Connected Components in the graph. The bridges represent vulnerabilities in a connected network, and they may be useful for designing and network reasoning. An example of a bridge is represented in Figure \ref{fig:FlowConstr1}a as the edge connecting  nodes 4 and 5. Note that this is the only edge in the graph whose suppression will affect the number of connected components.  

\begin{figure} 
    \centering
    \includegraphics[height=9cm,keepaspectratio]{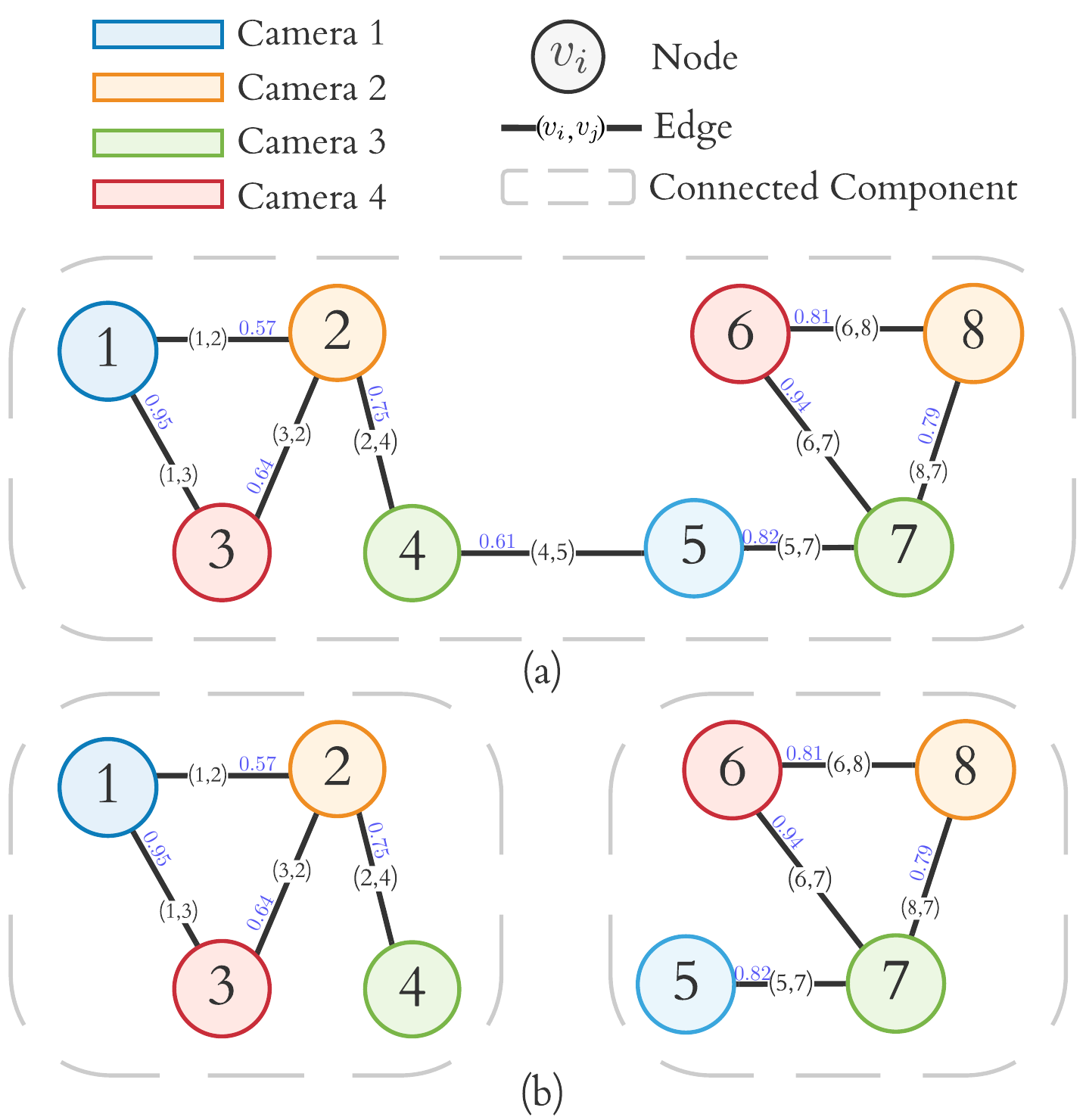}
    \caption{\textit{Splitting} Post-processing example of a graph with $M=4$. (a) shows the initial state of the graph after inference and before Post-processing stage. All nodes fulfill equation \ref{eq:3} condition. However, the graph is not fulfilling the condition in equation \ref{eq:4} since it is a unique connected component of size 8. First candidate edges to remove are the bridges (2,4), (4,5) and (5,7). (b) shows the final sub-graphs after removing the minimum probability bridge. \label{fig:FlowConstr2}}
\end{figure}

To ensure the flow constraint in Equation \ref{eq:3}, we perform the following \textit{prunning} strategy:

\begin{algorithm}[H]
\caption{\textit{Prunning}}\label{alg:alg1}
\begin{algorithmic}[1]
\STATE Compute the flow of each node in $V$

\STATE \textbf{while} there are nodes violating equation \ref{eq:3} \textbf{do}:
\STATE \hspace{0.5cm} \textbf{if} there is one bridge  \textbf{then}
\STATE \hspace{1cm} remove it  
\STATE \hspace{0.5cm} \textbf{if} there is more than one bridge  \textbf{then}
\STATE \hspace{1cm}  remove bridge edge with  $min(\hat{y}_{(v_i,v_j)}^L) $ 
\STATE \hspace{0.5cm} \textbf{if} there is no bridge  \textbf{then}
\STATE \hspace{1cm}  remove  edge with  $min(\hat{y}_{(v_i,v_j)}^L) $ 
\STATE \textbf{end}

\end{algorithmic}
\label{alg1}
\end{algorithm}



The second constraint that must be fulfilled refers to the size of each cluster. Since each identity should be represented by a unique connected component of nodes in the graph, it is logical to assume that
\begin{equation} \label{eq:4}
    |p_i| \leq M, 
\end{equation}
namely the maximum cardinality of each identity set should be, at most, the total number of cameras (see Figure \ref{fig:FlowConstr2}). To ensure the cardinality constraint we perform the following $splitting$ strategy:

\begin{algorithm}[H]
\caption{\textit{Splitting}}\label{alg:alg2}
\begin{algorithmic}[1]
\STATE Compute the size of each connected component in $G_t$
\STATE \textbf{while} there are components violating equation \ref{eq:4} \textbf{do}:
\STATE \hspace{0.5cm} \textbf{if} there is one bridge  \textbf{then}
\STATE \hspace{1cm} remove it  
\STATE \hspace{0.5cm} \textbf{if} there is more than one bridge  \textbf{then}
\STATE \hspace{1cm}  remove bridge edge with  $min(\hat{y}_{(v_i,v_j)}^L) $ 
\STATE \hspace{0.5cm} \textbf{if} there is no bridge  \textbf{then}
\STATE \hspace{1cm}  remove  edge with  $min(\hat{y}_{(v_i,v_j)}^L) $ 
\STATE \textbf{end}

\end{algorithmic}
\label{alg1}
\end{algorithm}


\subsection{Connected Components} \label{sec:CC}

Once the final graph is computed, to obtain the clusters of nodes containing different view detections of the same person identity, we compute the Connected Components (CCs) over the graph (see Figure \ref{fig:block-diagram}). A component, i.e., partition, of a graph is said to be a CC if there is a path between every pair of nodes.  Thus, each component will denote an identity, and each node in the component will correspond to a given identity in a certain camera view.

\section{Experiments}

\subsection{Evaluation framework}
\subsubsection{Datasets\label{sec:datasets}} 
 \begin{table}
 \renewcommand{\arraystretch}{1.2}
\caption{ \label{Tab:data-sets}Definition of the sets of data used for training and inference. Note that S1, S2 and S3 cover all possible combinations of EPFL sequences.}
\begin{center}

\resizebox{0.49\textwidth}{!}{
\begin{tabular}{cccc|ccc}
\hline 

 & \multicolumn{3}{c}{TRAINING} & \multicolumn{3}{c}{INFERENCE}\tabularnewline
\hline 

 & Laboratory & Terrace & Basketball  & Laboratory & Terrace & Basketball \tabularnewline
\hline \hline 
\textbf{S1} & \checkmark & \checkmark &  &  &  & \checkmark  \tabularnewline
\hline 
\textbf{S2} & \checkmark &  & \checkmark   &  & \checkmark &   \tabularnewline
\hline 
\textbf{S3} &  & \checkmark & \checkmark   & \checkmark &  &   \tabularnewline
\hline  


\end{tabular}}
\end{center}
\end{table}

\begin{figure} 
    \centering
    \includegraphics[width=0.49\textwidth,keepaspectratio]{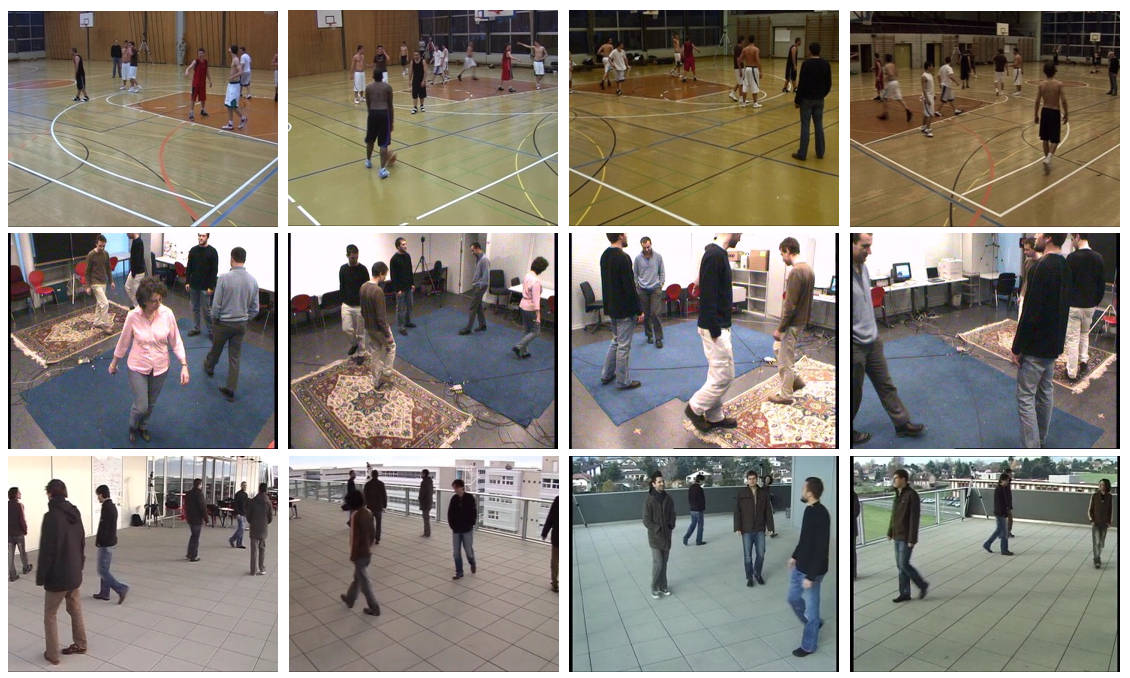}
    \caption{Sample frames from EPFL sequences: Basketball, Laboratory and Terrace. \label{fig:sample-views}}
\end{figure}

To evaluate the proposed approach, we require datasets that simultaneously provide multiple views of the same pedestrians and calibration information for camera-to-world mapping. We consider the EPFL\footnote{https://www.epfl.ch/labs/cvlab/data/data-pom-index-php/} multi-camera pedestrian videos dataset.
The EPFL dataset is divided in different sequences; we use the Terrace, Laboratory, and Basketball sequences, which are outdoor, indoor, and sport scenarios, respectively. All the sequences are captured by 4 cameras with overlapping FOVs and up to 9 pedestrians are present in them. Figure \ref{fig:sample-views} depicts sample frames from the 4 cameras of each sequence. Note that each sequence is totally independent from the others, in terms of scene location, lighting conditions, and the people who appear, so they correspond to totally different domains. 


Table \ref{Tab:data-sets} details the sub-sets of data we have defined for the experiments. Note that S1-3 sets are all possible training-inference combinations of EPFL sequences. We consider these combinations to prove that the method's operation does not depend on the visual appearance of the scene.

\subsubsection{Evaluation Metrics\label{sec:eval-metrics}} 
Since our objective is to group nodes in the graph corresponding to detections of the same person, we  evaluate our data association by measuring the clustering performance. 
We can evaluate the clustering performance using supervised metrics, as detections' identities are known.

In \cite{rosenberg2007v}, the following desirable objectives of cluster assignments were defined: Homogeneity: each cluster contains only samples of the same class; and Completeness: all samples of a given class are assigned to the same cluster. V-measure is also defined in \cite{rosenberg2007v} as the harmonic mean between Homogeneity and Completeness scores (as Precision and Recall are commonly combined into F-measure \cite{van1979information}). Homogeneity (H), Completeness (C) and V-measure (V-m) scores range from 0 to 1, where 1 represents perfect clustering.

These previous metrics are limited for the case of random labeling with a large number of clusters (i.e., scores may not be low). Therefore, we also consider metrics corrected-against-chance such as the Adjusted Rand Index (ARI) \cite{hubert1985comparing} that focuses on the assignment similarity, and the Adjusted Mutual Information (AMI) \cite{vinh2010information} that relies on the mutual information of two assignments  \cite{banerjee2005clustering}. Both ARI and AMI range from 1 (perfect match) to 0 (random labeling).

All the considered metrics (H, C, V-m, ARI and AMI) score in $[0,1]$, but we provide them in the interval $[0,100]$.

\subsubsection{Implementation Details}
As input of the proposed and compared approaches, we consider the ground-truth bounding boxes to avoid biases related to detector's performance. Furthermore, to remove dependency on the CNN feature extractor, we consider several of them publicly available in the state of the art. An ablation study on their influence is performed in Section \ref{sec:ab-feat}.  Each bounding box input image is resized according to the feature extractor employed, and also normalized by the mean and standard deviation of the ImageNet dataset \cite{russakovsky2015imagenet}. To reduce the overfitting and to improve generalization, we perform several random data augmentation techniques such as horizontal flip, color jitter and random erasing.

The proposed approach has been implemented using PyTorch and Pytorch Geometric frameworks, running on a NVIDIA TITAN RTX 24GB Graphics Processing Unit. To minimize the loss function and optimize the network parameters, we adopt the Stochastic Gradient Descend (SGD) solver.  Since our model is trained from scratch, and in order to avoid training instability, we perform, as proposed in \cite{goyal2017accurate}, a gradual warmup strategy that increases the learning rate from 0 to the initial learning rate linearly during 5 epochs.  The initial learning rate is heuristic set to $5e^{-3}$ and it decays along 20 epochs following the cosine annealing decay strategy, introduced in \cite{sgdr}, whose effectiveness with respect to the step decay has already been proven \cite{he2019bag,wightman2021resnet}. The batch size is set to 64, i.e., 64 graphs, since a graph per frame is computed (see Section \ref{sec:prob-form}).

In Table \ref{tab:sizes-fc} we report the specifications of each learnable encoder of the network, and the classifier.

\begin{table}
\begin{center}

\renewcommand{\arraystretch}{1.2}
\caption{Specification of each learnable encoder and the classifier of the network.  Each type, input and output sizes of each layer are detailed. Note that the input size of the layer 0 of $\mathcal{E}_v$ depends on the output dimension of the $CNN$ feature extractor (we consider ResNet50 as backbone).   \label{tab:sizes-fc}}
\resizebox{0.49\textwidth}{!}{

\begin{tabular}{ccccc}
\hline 
 & Layer & Type & Input & Output \tabularnewline
 \hline 
 \multicolumn{5}{c}{ \textbf{Feature Encoders}} \tabularnewline
 \hline 
\multirow{2}{*}{\rotatebox{90}{$\mathcal{E}_v$}} & 0  & FC + ReLU &  512 & 128\tabularnewline
\cline{2-5}
& 1 & FC + ReLU &  128 & 32\tabularnewline
\hline
\rotatebox{90}{$\mathcal{E}_e$} & 0  & FC + ReLU &  4 & 6 \tabularnewline
\hline 

\multicolumn{5}{c}{\textbf{Message Passing Nework}} \tabularnewline
\hline \hline
\rotatebox{90}{$\mathcal{U}_v$} & 0  & FC + ReLU &  38 & 32 \tabularnewline
\hline
\rotatebox{90}{$\mathcal{U}_e$} & 0  & FC + ReLU &  70 & 6 \tabularnewline
\hline
\multicolumn{5}{c}{\textbf{Classifier}} \tabularnewline
\hline \hline
\multirow{2}{*}{\rotatebox{90}{$\mathcal{C}$}} & 0  & FC + ReLU &  6 & 4\tabularnewline
\cline{2-5}
& 1 & FC + Sigmoid &  4 & 1\tabularnewline
\hline

\end{tabular}}
\end{center}
\end{table}

\subsection{Ablation Study}
We present an ablation study to analyze the performance impact of parameters and strategies considered in the proposed approach. Firstly, the influence of different appearance descriptors is evaluated. Secondly, the effect of the number of message passing steps is analysed, and, finally, the post-processing strategies are assesed.

\subsubsection{Appearance descriptor \label{sec:ab-feat}}

Table \ref{tab:feat-cnn} shows the algorithm performance when varying the feature extractor used to obtain the appearance descriptor, that is used to initialize the nodes and edges embeddings in the graph. For the sake of simplicity and space, we perform this comparison over the S1 set; similar results are obtained for the other sets. 

In \cite{braso2020learning}, a model using ResNet50 as backbone and trained on Market1501 \cite{zheng2015scalable}, CUHK03 \cite{li2014deepreid}, and DukeMTMC \cite{ristani2016performance} datasets was provided. A new backbone, Omni-Scale Network (OSNet), was introduced in \cite{zhou2019omni}. We employ their same-domain model, trained in Market1501, and a multi-source model, introduced by \cite{zhou2021learning}, and trained on MSMT17 \cite{wei2018person}, CUHK03, and DukeMTMT. Lastly, \cite{quispe2021top}  proposed the Top-DB-Net architecture, using ResNet50 as baseline. They only provide same-domain models, thus we consider the ones trained on CUHK03 and in Market1501. Table \ref{tab:feat-cnn} shows the comparison in terms of clustering evaluation metrics obtained with all the previously mentioned models; in addition, the first row shows the performance considering ResNet50 as backbone just pre-trained on the Imagenet \cite{russakovsky2015imagenet} dataset. 

Intuitively, we expect that a better descriptor yields a better performance. The results of this experiment seem to prove this hypothesis, since using a trained descriptor devoted to pedestrian re-identification increases the performance, at least, a 35.33\% in terms of ARI, a 27.02\% regarding ARI and 12.76\% of V-m. 
Note that each ReID model is trained with a different image size. Since a bigger image size implies more computational resources, we believe, in the light of the results, that using the model trained by \cite{braso2020learning}, second row of the table, is a good trade-off between image size and performance, thus we adopt this model for the following ablation experiments.

\begin{table}
\begin{center}

\renewcommand{\arraystretch}{1.2}
\caption{Ablation study on CNN REID appearance model. Datasets key: I = Imagenet; M = Market-1501; C = CUHK03; D =  DukeMTMC; MS = MSMT17. \label{tab:feat-cnn}}
\resizebox{0.49\textwidth}{!}{

\begin{tabular}{ccccc|ccccc}
\hline 

 & \multicolumn{4}{c}{Feature Extractor} & \multicolumn{5}{c}{Clustering evaluation metrics}\tabularnewline
\hline 
 & Method & Backbone & Source & Size & ARI & AMI & H & C & V-m\tabularnewline
\hline \hline 

\multirow{6}{*}{S1} & - & ResNet 50 & I & {[}128, 64{]} & 49.37 & 61.26 & 71.63 & 89.41 & 75.77\tabularnewline
\cline{2-10}
 & \cite{braso2020learning} & ResNet 50 & M + C + D & {[}128, 64{]} & 70.23 & 80.21 & 81.74 & 95.63 & 87.01\tabularnewline \cline{2-10}

 & \cite{zhou2019omni} & OSNet  & M & {[}256, 128{]} & 74.24 & 82.38 & 85.35 & 94.44 & 88.99\tabularnewline \cline{2-10}

 & \cite{zhou2021learning} & OSNet & MS + C + D & {[}256, 128{]} & 73.10 & 76.47 & 64.20 & 83.93 & 87.91\tabularnewline \cline{2-10}

 & \cite{quispe2021top} & Top-DB-Net & C & {[}384, 128{]} & 66.81 & 77.81 & 79.41 & 95.50 & 85.44\tabularnewline\cline{2-10}

 & \cite{quispe2021top} & Top-DB-Net & M & {[}384, 128{]} & 79.54 & 85.79 & 88.01 & 95.39 & 90.97\tabularnewline
\hline
\end{tabular}}
\end{center}
\end{table}

\subsubsection{Number of message passing steps}
This experiment studies the impact of performing different number of message passing steps or iterations ($L$). It could be reasonable to expect that higher values of $L$ yield better performance, since the nodes and edges information is spread farther along the graph, in other words, more neighbours features is involved in the edge prediction.

\begin{figure} 
    \centering
    
    \includegraphics[width=0.49\textwidth,keepaspectratio]{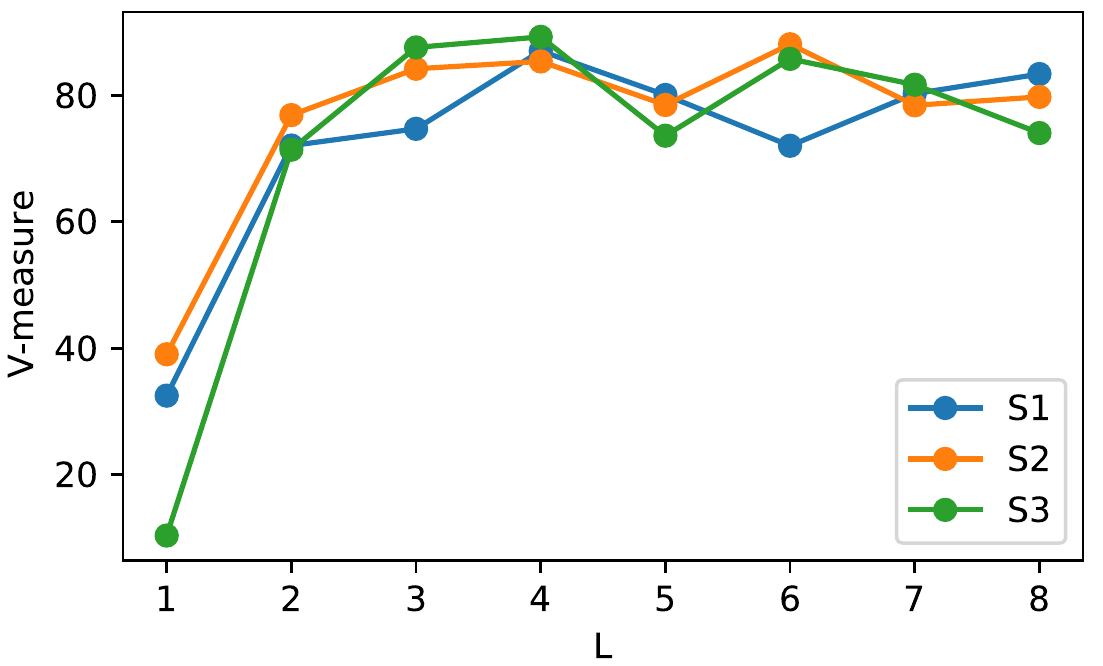}
    \caption{Ablation study on the effect of the number of message passing steps. \label{fig:L-ablation}}
\end{figure}

We train the network with $L=[1,8]$ and we report V-measure metric, capturing both Homogeneity and Completeness, in Figure \ref{fig:L-ablation}. For reliable results, we  perform this experiment over the three sets of data. As expected, an upward tendency is observed. We can also note that the biggest rise occurs when instead of just looking at the first adjacent neighbour, higher-order information is taken into account. Lastly, we note the performance is slightly stabilized for $L\ge 3$, i.e.,  when each node knows about neighbours at a distance of 3 and forward.   The topology of the scenario under consideration, ($M=4$ cameras), may be a reason for this behaviour.

In the light of the results, we consider $L=4$, as a peek of performance, for the further experiments.

 \subsubsection{Post-processing strategies} 
 As described in section \ref{seq:post}, \textit{pruning} forces each node to be connected to, at most, $M-1$ nodes, thus each pedestrian identity can be composed of, at most, detections of $M$ different views. The \textit{splitting} strategy ensures pedestrians identity clusters to be at most of $M$ size. 
 
Table \ref{tab:post-strategies} shows the clustering evaluation metrics, previously detailed in section \ref{sec:eval-metrics}; of the proposed data association algorithm when performing no post-processing, each post-processing strategy independently, and both in conjunction. We list the inference performance of all the sets S1-3 and the average of them.  From Table \ref{tab:post-strategies}, we can observe that both strategies independently increase the global association performance, measured by ARI and AMI indexes, and in conjunction, both outperform the results. Focusing on the average performance: \textit{Pruning}, individually, yields a performance increment of 20.7\% and 12.7\%, for both indexes respectively; \textit{splitting} increases ARI and AMI by 19.3\% and 10.74\%; whilst, both together report increases of 30.9\% and 16.1\%.  The  Homogeneity is considerably improved with both strategies, since they inherently break wrong large clusters into smaller ones, yielding an increase in H of 25.7\%. Consequently, a 4.89\% is sacrificed regarding Completeness, since,  it is easier for all the samples of a class to belong to the same cluster if there are artificially large clusters. Note that the best V-measure is reached when combining both strategies.

 \begin{table}
 \begin{center}
  \renewcommand{\arraystretch}{1.2}

\caption{Ablation study on the effect of \textit{pruning} and \textit{splitting} post-processing strategies. Average results are underlined and the best average results are in bold. \label{tab:post-strategies}}
\resizebox{0.49\textwidth}{!}{

\begin{tabular}{ccc|ccccc}
\hline 

 & \multicolumn{2}{c}{Post-processing} & \multicolumn{5}{c}{Clustering evaluation metrics}\tabularnewline
\hline 
 & Pruning & Splitting & ARI & AMI & H & C & V-m\tabularnewline
\hline \hline 
S1 &  &  & 70.23 & 80.21 & 81.74 & 95.63 & 87.01\tabularnewline
\hline 
S2 &  &  & 69.68 & 80.03 & 80.08 & 96.48 & 85.29\tabularnewline
\hline 
S3 &  &  & 54.32 & 65.96 & 61.92 & 99.78 & 70.71\tabularnewline
\hline 
avg. &  &  & \underline{64.74} & \underline{75.46} & \underline{74.58} & \underline{97.29} & \underline{81.00}\tabularnewline
\hline 
\hline 
S1 & \checkmark &  & 78,07 & 84,33 & 88,21 & 93,82 & 90,39\tabularnewline
\hline 
S2 & \checkmark &  & 77.90 & 85.27 & 87.96 & 94.70 & 90.68\tabularnewline
\hline 
S3 & \checkmark &  & 78.46 & 85.67 & 84.20 & 98.31 & 89.17\tabularnewline
\hline 
avg. & \checkmark &  & \underline{78.14} & \underline{85.09} & \underline{85.56} & \underline{95.61} & \underline{90.08}\tabularnewline
\hline 
\hline 
S1 &  & \checkmark & 77,45 & 82,24 & 89,73 & 90,47 & 89,80\tabularnewline
\hline 
S2 &  & \checkmark & 76.88 & 83.62 & 87.86 & 92.98 & 89.43\tabularnewline
\hline 
S3 &  & \checkmark & 77.43 & 84.96 & 84.70 & 96.59 & 89.36\tabularnewline
\hline 
avg. &  & \checkmark & \underline{77.25} & \underline{83.60} & \underline{87.51} & \underline{93.34} & \underline{89.53}\tabularnewline
\hline 
\hline 
S1 & \checkmark & \checkmark & 82.99 & 85.12 & 94.23 & 89.97 & 91.94\tabularnewline
\hline 
S2 & \checkmark & \checkmark & 83.07 & 86.77 & 93.59 & 92.03 & 92.66\tabularnewline
\hline 
S3 & \checkmark & \checkmark & 88.24 & 91.07 & 93.63 & 95.59 & 94.42\tabularnewline
\hline 
avg. & \checkmark & \checkmark & \textbf{\underline{84.76}} & \textbf{\underline{87.65}} & \textbf{\underline{93.81}} & \textbf{\underline{92.53}} & \textbf{\underline{93.00}}\tabularnewline
\hline 
\end{tabular}}
\end{center}
\end{table}

\subsection{Comparison with the state-of-the-art}

\begin{table*}
 \renewcommand{\arraystretch}{1.2}
\caption{Comparison with the state of the art association techniques, considering ResNet50 trained on Market-1051, CUHK03 and DukeMTMC datasets as feature extractor. Average results are underlined, and the best average results are in bold. \label{Tab:results-soa-resnet} }
\begin{center}

\resizebox{0.99\textwidth}{!}{
\begin{tabular}{cccccccccc}
\hline 
\multirow{2}{*}{} & \multirow{2}{*}{\raggedright{} \makecell{Cross-camera \\ Data Association}} & \multicolumn{3}{c}{Feature Extractor} & \multirow{2}{*}{Rand Index} & \multirow{2}{*}{Mutual Information} & \multirow{2}{*}{Homogeneity} & \multirow{2}{*}{Completeness} & \multirow{2}{*}{V-measure}\tabularnewline
\cline{3-5}
 &  & Method & Backbone & Source &  &  &  &  & \tabularnewline
\hline \hline 
S1 &\multirow{4}{*}{\makecell{Apperance-based \\ (L2 + th) \\ \cite{Narayan2017,ristani2018features,he2020multi}}} & \multirow{4}{*}{\cite{braso2020learning}} & \multirow{4}{*}{ResNet50} & \multirow{4}{*}{M + C + D} & 17.42 & 25.24 & 31.57 & 76.41 & 39.79\tabularnewline
S2 &  &  &  &  & 12.09 & 15.61 & 26.87 & 66.78 & 28.10\tabularnewline
S3 &  &  &  &  & 54.54 & 62.32 & 71.17 & 82.77 & 73.54\tabularnewline
avg &&  &  &  & \underline{28.01} & \underline{34.39} & \underline{43.20} & \underline{75.43} & \underline{47.14}\tabularnewline
\hline 
S1 &\multirow{4}{*}{\makecell{Apperance-based \\ (Cosine dist. +  th) \\ \cite{Maksai_2019_CVPR,zhou2021learning}}}  & \multirow{4}{*}{\cite{braso2020learning}} & \multirow{4}{*}{ResNet50} & \multirow{4}{*}{M + C + D} & 15.52 & 22.93 & 28.48 & 77.74 & 36.54\tabularnewline
S2 & &  &  &  & 12.34 & 15.87 & 27.71 & 66.06 & 29.40\tabularnewline
S3 &  &  &  &  & 54.16 & 62.03 & 71.38 & 82.25 & 73.59\tabularnewline
avg. & &  &  &  & \underline{27.34} & \underline{33.61} & \underline{42.52} & \underline{75.35} & \underline{46.21}\tabularnewline
\hline 
S1 & \multirow{4}{*}{\makecell{ReID \\ (Top-1 rank)}} & \multirow{4}{*}{\cite{braso2020learning}} & \multirow{4}{*}{ResNet50} & \multirow{4}{*}{M + C + D} & 35.82 & 44.021 & 57.15 & 71.17 & 61.89\tabularnewline
S2 &  &  &  &  & 30.42 & 38.17 & 60.27 & 32.52 & 57.48\tabularnewline
S3 &  &  &  &  & 66.64 & 71.67 & 79.93 & 84.20 & 80.74\tabularnewline
avg &  &  &  &  & \underline{44.29} & \underline{51.28} & \underline{65.78} & \underline{63.63} & \underline{66.70}\tabularnewline

\hline 
S1 & \multirow{4}{*}{\makecell{Geometrical approach \\  \cite{lopez2018semantic}}} & \multirow{4}{*}{-} & \multirow{4}{*}{-} & \multirow{4}{*}{-} & 54.93 & 71.03 & 68.44 & 98.76 & 79.20\tabularnewline
S2 &  &  &  &  & 30.72 & 40.34 & 37.37 & 99.95 & 44.36\tabularnewline
S3 &  &  &  &  & 44.83 &54.84  &51.63  & 99.21 & 59.33\tabularnewline
avg &  &  &  &  & \underline{43.49} & \underline{55.40} & \underline{52.48} & \underline{99.03} & \underline{60.96} \tabularnewline

\hline 
S1 & \multirow{4}{*}{\makecell{Geometrical approach \\ + Appearance (L2 + th) \cite{lima2021generalizable}}} & \multirow{4}{*}{\cite{braso2020learning}} & \multirow{4}{*}{ResNet50} & \multirow{4}{*}{M + C + D} & 57.66 & 68.10 & 80.11 & 85.07 & 81.61\tabularnewline
S2 &  &  &  &  &40.53 & 50.96 & 68.89 & 76.12 & 66.72 \tabularnewline
S3 &  &  &  &  & 73.47 & 77.50 & 91.07 & 84.95  & 87.02  \tabularnewline
avg &  &  &  &  & \underline{57.22} & \underline{65.52} & \underline{80.02} & \underline{82.04} & \underline{78.45} \tabularnewline
\hline 


S1 & \multirow{4}{*}{\makecell{GNN-CCA (ours)}} & \multirow{4}{*}{\cite{braso2020learning}} & \multirow{4}{*}{ResNet50} & \multirow{4}{*}{M + C + D} & 82.99 & 85.12 & 94.23 & 89.97 & 91.94\tabularnewline
S2 &  &  &  &  & 83.07 & 86.77 & 93.59 & 92.03 & 92.66\tabularnewline
S3 &  &  &  &  & 88.24 & 91.07 & 93.63 & 95.59 & 94.42\tabularnewline
avg &  &  &  &  & \textbf{\underline{84.76}} & \textbf{\underline{87.65}} & \textbf{\underline{93.81}} & \textbf{\underline{92.53}} & \textbf{\underline{93.00}}\tabularnewline
\hline 

\end{tabular}}
\end{center}
\end{table*}

 This section compares our proposal, let us denote it GNN-CCA (GNN for Cross-Camera data Association),  with state-of-the-art cross-camera data association approaches.
We report in Table \ref{Tab:results-soa-resnet} the results for each set of data S1-3, and also the average of them, considering as feature extractor the ResNet50 backbone trained by \cite{braso2020learning}, since it has proven to be a good trade-of between computational cost and performance. GNN-CCA is compared with the following algorithms.

\textbf{Appearance-based (L2/Cosine dist. + th).} We measure the efficiency of the widespread use of the Euclidean (L2) \cite{Narayan2017,ristani2018features,he2020multi} and Cosine \cite{Maksai_2019_CVPR,zhou2021learning} distances between appearance descriptors to associate cross-camera detections. In this case, distances also need to be thresholded. We compute the optimal threshold for each set of data S1-3 to perform a fair comparison. For this aim, we normalize all the appearance distances of each set within the interval [0,1] (by dividing by the maximum distance). Then, we evaluate the following range of thresholds: $th= \left\{0.1 \cdot k,\; k \in [0, 10] \right\}$ and the best performance is reported. The results in Table \ref{Tab:results-soa-resnet} are obtained with the following optimal thresholds: 0.63, 0.59, and 0.55 for, respectively, S1, S2, and S3 when using the L2 distance; and 0.71, 0.71, and 0.75 when considering the Cosine distance. Therefore, pairs of nodes whose distances between descriptors are under these thresholds will be connected.

\textbf{ReID (Top-1 Rank).} The task of image-based people re-identification aims to match a probe person (typically known as a query image) with a gallery set of images, looking for an identity correspondence, by producing a ranked list of all the gallery images based on their appearance similarity. The next approach for comparison estimates how the re-identification ranking technique performs for the task at hand.
We treat each node in the graph as a query and the rest as the gallery, which is ranked in decreasing order based on their Euclidean distances between descriptors \cite{dai2019batch,quispe2021top}. We report the performance when connecting each node to its top-1 rank node.

\textbf{Geometrical approach.} We have implemented the geometrical approach proposed in \cite{lopez2018semantic}, which considers traditional graphs where nodes representing pedestrians are connected based on their ground plane spatial location. As the authors propose, two nodes in the graph are connected if their Euclidean spatial distance is under a threshold. We consider the original setting, since they also report results using the EPFL benchmark.

\textbf{Geometrical approach + Appearance (L2 + th).} We also have implemented the extended version of the geometrical approach, proposed by \cite{lima2021generalizable}. This approach adds a new condition to the graphs construction, based on the distance between appearance features. Now, two nodes are connected according to their thresholded Euclidean spatial and appearance distances. Again, for a fair comparison, we consider the optimal computed threshold for each set of data. 

The results in Table \ref{Tab:results-soa-resnet} show a considerable increment of performance of GNN-CCA over the rest of approaches. Regarding the V-measure, a relative percentage increment of  the 97.28\% is obtained with respect to the commonly used L2 distance for comparing appearance features. An increment of the 18.55\% is obtained with respect to the best approach, the combination of geometrical and appearance features. Both percentage are measured considering the average V-m of the tree sets S1-3. As might be expected, putting some effort into jointly learning features, and a metric that is tailored to the features and data, instead of considering non-learnable pre-defined distances, improves considerably the performance. Our method represents a valuable alternative to the techniques currently used to solve the task of associating cross-camera pedestrian detections in bigger pipelines. Furthermore,  our training proposal only has around 0.2M parameters, which makes it a very light network easy to train (for reference, ResNet50 has 12M). Moreover, as each sequence is independent from each other (see Section \ref{sec:datasets}),  we can verify that our proposal is totally able to generalize for different domains, and it does not require to be trained in the same scenario in which it is tested. 
We also report the performance when using Top-DB-Net as feature extractor (see Table \ref{Tab:results-soa-topdb}). The same trend regarding the performance evaluation can be observed, thus, it suggest that our approach is also modular, not dependent on a particular feature extractor algorithm.

\begin{table*}
 \renewcommand{\arraystretch}{1.2}
\caption{Comparison with the state of the art association techniques, considering Top-DB-Net trained on Market-1051 as feature extractor. Average results are underlined, and the best average results are in bold.\label{Tab:results-soa-topdb} }
\begin{center}

\resizebox{0.99\textwidth}{!}{
\begin{tabular}{cccccccccc}
\hline 
\multirow{2}{*}{} & \multirow{2}{*}{\raggedright{}Cross-camera Data Association} & \multicolumn{3}{c}{Feature Extractor} & \multirow{2}{*}{Rand Index} & \multirow{2}{*}{Mutual Information} & \multirow{2}{*}{Homogeneity} & \multirow{2}{*}{Completeness} & \multirow{2}{*}{V-measure}\tabularnewline
\cline{3-5}
 &  & Method & Backbone & Source &  &  &  &  & \tabularnewline
\hline \hline

S1 & \multirow{4}{*}{\makecell{Apperance-based \\ (L2 + th) \\ \cite{Narayan2017,ristani2018features,he2020multi}}} & \multirow{4}{*}{\cite{quispe2021top}} & \multirow{4}{*}{Top-DB-Net} & \multirow{4}{*}{M} & 4.22 & 6.35 & 16.80 & 64.66 & 23.22\tabularnewline
S2 & &  &  &  & 4.21 & 5.97 & 31.26 & 47.07 & 26.35\tabularnewline
S3 & &  &  &  & 54.80 & 61.40 & 73.40 & 80.94 & 74.29\tabularnewline
avg. & &  &  &  & \underline{21.07} & \underline{24.57} & \underline{40.48} & \underline{64.22} & \underline{41.28}\tabularnewline
\hline 

S1 & \multirow{4}{*}{\makecell{Apperance-based \\ (Cosine dist. +  th) \\ \cite{Maksai_2019_CVPR,zhou2021learning}}} & \multirow{4}{*}{\cite{quispe2021top}} & \multirow{4}{*}{Top-DB-Net} & \multirow{4}{*}{M} & 1.56 & 2.01 & 10.05 & 65.20 & 15.16\tabularnewline
S2 &  &  &  &  & 3.40 & 4.34 & 2.22 & 51.64 & 18.16\tabularnewline
S3 &  &  &  &  & 52.17 & 59.47 & 68.39 & 81.62 & 71.19\tabularnewline
avg. &  &  &  &  & \underline{19.04} & \underline{21.94} & \underline{26.88} & \underline{63.15} & \underline{34.89}\tabularnewline

\hline 
S1 & \multirow{4}{*}{\makecell{ReID \\ (Top-1 rank)}} & \multirow{4}{*}{\cite{quispe2021top}} & \multirow{4}{*}{Top-DB-Net} & \multirow{4}{*}{M} & 21.45 & 28.03 & 44.61 & 60.66 & 49.25\tabularnewline
S2 &  &  &  &  & 18.90 & 25.22 & 45.55 & 55.47 & 44.89\tabularnewline
S3 &  &  &  &  & 60.97 & 66.63 & 76.88 & 80.99 & 76.94\tabularnewline
avg &  &  &  &  & \underline{33.77} & \underline{39.96} & \underline{55.68} & \underline{65.70} & \underline{57.02}\tabularnewline
\hline 
S1 & \multirow{4}{*}{\makecell{Geometrical approach \\  \cite{lopez2018semantic}}} & \multirow{4}{*}{-} & \multirow{4}{*}{-} & \multirow{4}{*}{-} & 54.93 & 71.03 & 68.44 & 98.76 & 79.20\tabularnewline
S2 &  &  &  &  & 30.72 & 40.34 & 37.37 & 99.95 & 44.36\tabularnewline
S3 &  &  &  &  & 44.83 &54.84  &51.63  & 99.21 & 59.33\tabularnewline
avg &  &  &  &  & \underline{43.49} & \underline{55.40} & \underline{52.48} & \underline{99.03} & \underline{60.96} \tabularnewline
\hline

S1 & \multirow{4}{*}{\makecell{Geometrical approach + \\ Appearance (L2 + th) \cite{lima2021generalizable}}} & \multirow{4}{*}{\cite{quispe2021top}} & \multirow{4}{*}{Top-DB-Net} & \multirow{4}{*}{M } & 50.09 & 60.52 & 77.32 & 80.84 & 77.91 \tabularnewline
S2 & &  & & & 28.75 & 38.38 & 70.60  & 63.39  & 60.21 \tabularnewline
S3 &  &  &  &  & 70.59 &74.14 & 91.35 & 82.82 & 86.03 \tabularnewline
avg &  &  &  &  & \underline{49.81} & \underline{57.68} & \underline{79.75} & \underline{75.68} & \underline{74.71} \tabularnewline
\hline


S1 & \multirow{4}{*}{\makecell{GNN-CCA (ours)}} & \multirow{4}{*}{\cite{quispe2021top}} & \multirow{4}{*}{Top-DB-Net} & \multirow{4}{*}{M} & 96.22 & 87.95 & 95.18 & 91.71 & 93.31\tabularnewline
S2 &  &  &  &  & 32.35 &33.26  &95.54  &69.08 & 78.92\tabularnewline
S3 &  &  &  &  & 62.51 & 66.48 & 87.06 & 79.79 & 82.36 \tabularnewline
avg &  & &  &  & \textbf{\underline{63.69}}& \textbf{\underline{62.56}} & \textbf{\underline{92.59}} & \textbf{\underline{80.19}} & \textbf{\underline{84.86}}\tabularnewline
\hline 
\end{tabular}}
\end{center}
\end{table*}

\section{Conclusions}
Cross-camera data association is a crucial task in multiple multi-camera computer vision fields. We propose, to the best of our knowledge, the first learnable approach fully devoted to tackling it, by jointly learning both the feature representations and the similarity metric. To this aim, we consider Graph Neural Networks, that have been previously unused in the cross-camera scope, and that have already proven its performance in other single-camera scopes. The proposed design can be applied for cross-camera matching of any type of objects, although we validate it in pedestrians scenarios. The results show that the proposal outperforms the state-of-the-art techniques currently used to solve the task of associating cross-camera detections. Furthermore, a global association is made, being an advantage over pairwise alternatives, maintaining  a low computational complexity. The proposed method represents an effective and lightweight technique for the association of cross-camera detections, with a performance improvement over other alternatives that benefits the overall results of the target multi-camera computer vision task.

\section*{Acknowledgments}
This work was partially supported by the Spanish Government (TEC2017-88169-R MobiNetVideo).


\bibliographystyle{IEEEtran}
\bibliography{references.bib}

\section{Biography Section}
\begin{IEEEbiography}[{\includegraphics[width=1in,height=1.25in,clip,keepaspectratio]{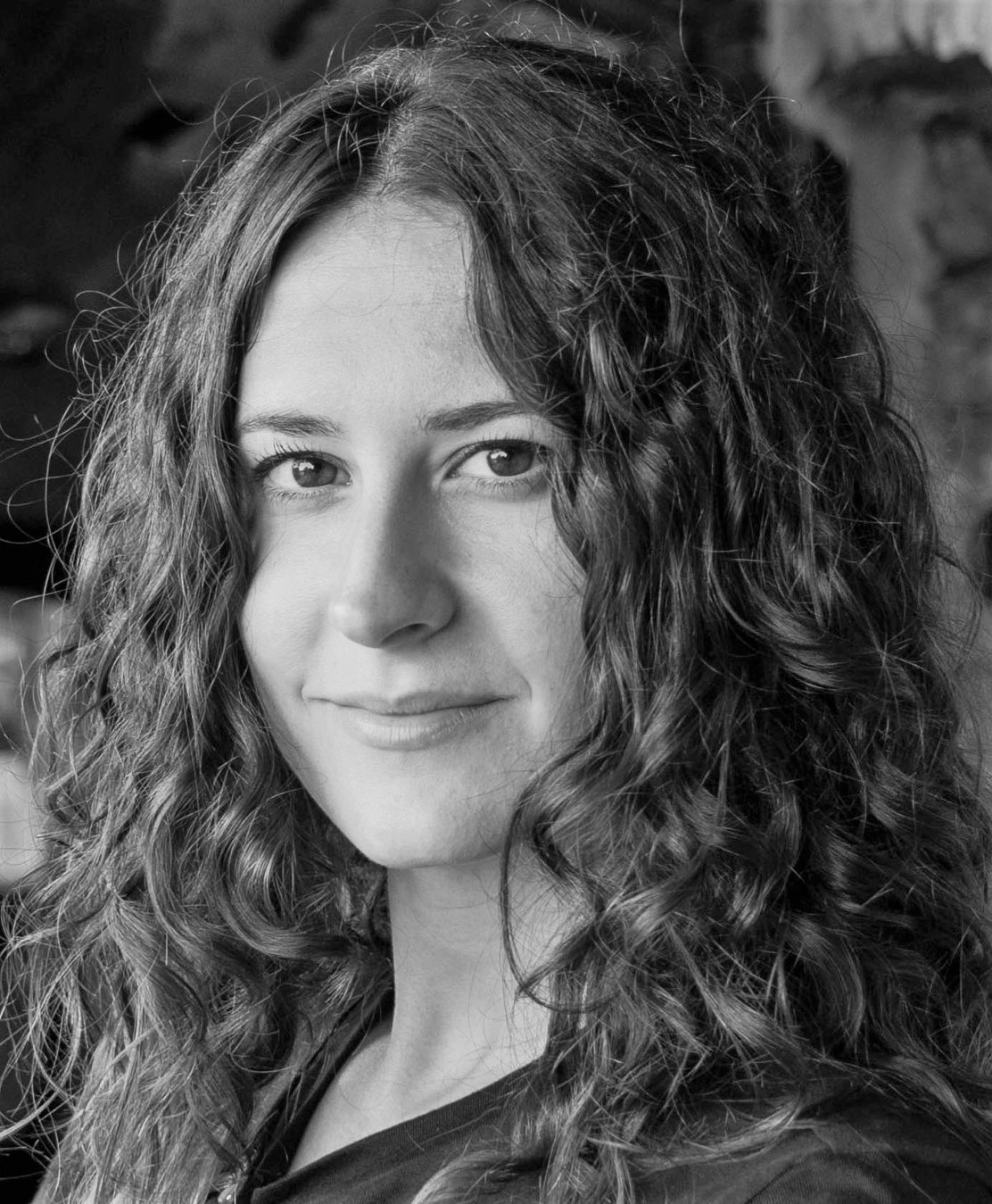}}]{Elena Luna García}

obtained a B.S degree in Telecommunications Engineering in 2015 at the Universidad Autónoma de Madrid (Spain). In 2017 she received the M.S. degrees belonging to the International Joint Master Program in Image Processing and Computer Vision (IPCV) at the PPCU in Budapest (Hungary), the University of Bordeaux (France) and the UAM (Spain). She is currently pursuing the Ph.D. degree with the Video Processing and Understanding Lab (VPU-Lab) at the UAM (Spain).

\end{IEEEbiography}

\vspace{11pt}

\begin{IEEEbiography}[{\includegraphics[width=1in,height=1.25in,clip,keepaspectratio]{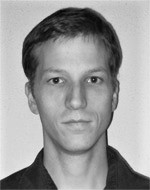}}]{Juan C. SanMiguel}
received the Ph.D. degree
in computer science and telecommunication from
University Autonoma of Madrid, Madrid, Spain,
in 2011. He was a Post-Doctoral Researcher with
Queen Mary University of London, London, U.K.,
from 2013 to 2014, under a Marie Curie IAPP
Fellowship. He is currently Associate Professor at
University Autónoma of Madrid and Researcher
with the Video Processing and Understanding Laboratory. His research interests include computer vision
with a focus on online performance evaluation and
multicamera activity understanding for video segmentation and tracking. He
has authored over 40 journal and conference papers.
\end{IEEEbiography}
\vspace{11pt}

\begin{IEEEbiography}[{\includegraphics[width=1in,height=1.25in,clip,keepaspectratio]{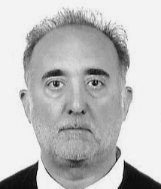}}]{ José M. Martínez}received the Ph.D. degree in
computer science and telecommunication from the
Universidad Politécnica de Madrid, Madrid, Spain,
in 1998. He is currently a Full Professor with the
Escuela Politécnica Superior, Universidad Autónoma
de Madrid, Madrid. He has acted as an auditor and
a reviewer for the EC for projects of the frameworks
program for research in Information Society and
Technology (IST). He is the author or coauthor of
more than 100 papers in international journals and
conferences and a coauthor of the first book about
the MPEG-7 standard published in 2002. His professional interests cover
different aspects of advanced video surveillance systems and multimedia
information systems.

\end{IEEEbiography}

\begin{IEEEbiography} [{\includegraphics[width=1in,height=1.25in,clip,keepaspectratio]{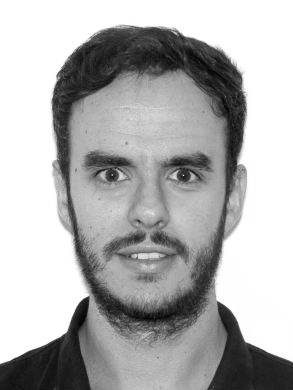}}]
{Pablo Carballeira} received the Telecommunication Engineering degree
(five years engineering program), Communications Technologies and
Systems Master degree (two years MS program) and the Ph.D. degree in
Telecommunication from the Universidad Polit\'ecnica de Madrid (UPM)
in 2007, 2010 and 2014 respectively. From 2008 to 2017 he has been
a member of the Grupo de Tratamiento de Im\'agenes (Image Processing
Group) at the UPM. Since 2017 he is an Assistant Professor, and a
member of the Video Processing and Understanding Lab, at the Universidad
Aut\'onoma de Madrid (UAM). His research interests include 
computer vision, video coding, and quality of experience evaluation for immersive
visual media. He has been actively involved in European projects, national projects, and standardization activities from ISO\textquotesingle s Moving Picture
Experts Group (MPEG) related to lightfield and free-navigation video
technologies.
\end{IEEEbiography}

\vfill

\end{document}